\documentclass[graybox]{svmult}

\usepackage[numbers]{natbib}

\usepackage{mathptmx}       
\usepackage{afterpage}       

\usepackage{helvet}         
\usepackage{courier}        
\usepackage{type1cm}        
\usepackage{makeidx}         
\usepackage{graphicx}        
\usepackage{multicol}        
\usepackage[bottom]{footmisc}
\usepackage{amsmath}
\usepackage{bm}
\usepackage{amssymb}
\usepackage{gensymb}
\usepackage{siunitx} 
\usepackage{algorithm}
\usepackage{algpseudocode}
\usepackage{tikz}
\usepackage{listings}
\usepackage{multirow}
\usepackage{microtype}
\usepackage{cleveref}

\makeindex             

\begin{document}

\title*{Super-Resolution of PROBA-V Images Using Convolutional Neural Networks}
\titlerunning{Super-Resolution of PROBA-V}
\author{Marcus M\"{a}rtens \and Dario Izzo \and Andrej Krzic \and Dani\"el Cox}
\institute{Dario Izzo \at European Space Agency, Noordwijk, 2201 AZ, The Netherlands\\ \email{dario.izzo@esa.int}
\and Marcus M\"{a}rtens \at European Space Agency, Noordwijk, 2201 AZ, The Netherlands\\ \email{marcus.maertens@esa.int} 
\and  Andrej Krzic  \at European Space Agency, Noordwijk, 2201 AZ, The Netherlands\\ \email{Andrej.Krzic@iof.fraunhofer.de}
\and Dani\"el Cox \at European Space Agency, Noordwijk, 2201 AZ, The Netherlands\\ \email{d.w.s.cox@alumnus.utwente.nl}
}
%
%
\maketitle

\abstract{ESA's PROBA-V Earth observation satellite enables us to monitor our planet at a large scale, studying the interaction between vegetation and climate and provides guidance for important decisions on our common global future. 
However, the interval at which high resolution images are recorded spans over several days, in contrast to the availability of lower resolution images which is often daily. 
We collect an extensive dataset of both, high and low resolution images taken by PROBA-V instruments during monthly periods to investigate Multi Image Super-resolution, a technique to merge several low resolution images to one image of higher quality.
We propose a convolutional neural network that is able to cope with changes in illumination, cloud coverage and landscape features which are challenges introduced by the fact that the different images are taken over successive satellite passages over the same region.
Given a bicubic upscaling of low resolution images taken under optimal conditions, we find the Peak Signal to Noise Ratio of the reconstructed image of the network to be higher for a large majority of different scenes.
This shows that applied machine learning has the potential to enhance large amounts of previously collected earth observation data during multiple satellite passes.}

\section{Introduction}
\label{sec:introduction}

Techniques for improving the resolution of images by digital means are commonly referred to as Super Resolution (SR) image reconstruction techniques and are comprehensively reviewed by Park et al.~\citep{park2003super} and Nasrollahi et al.~\cite{Nasrollahi2014Super}. 
These techniques can be split into two main categories: \emph{Single image Super-resolution} (SISR) and \emph{Multi image Super-resolution} (MISR).
The latter makes use of multiple images of the same scene in order to create one single higher resolution image and it is of obvious interest for earth observation (EO) satellites, where the availability of multiple images of the same location is common.
Some of these satellites, such as SPOT-VGT \citep{latry2009staggered} or ZY-3 TLC \citep{li2017super}, have payloads that are able to take multiple images of the same area during a single satellite pass, which creates a most favorable condition for the use of SR algorithms as images are taken simultaneously.
Thus, the SR can be already included in the final product of these satellites during post-processing on the ground. Moreover, there are also multiple images available from successive revisits of a satellite over the same coordinates.
In this scenario, the images might differ quite significantly due to the larger scales of time involved (hours or days in comparison to seconds). Several, detrimental effects for the performances of SR algorithms might be introduced: difference in illumination conditions, difference in cloud coverage and even differences in actual pixel content.
The question arises, whether under this conditions MISR techniques still apply and if so, under which conditions they may provide an increase in image quality?

In this work we document the collection of a large-scale data-set of images from the European Space Agency's (ESA) vegetation observation satellite PROBA-V \citep{proba} for the purpose to study post-acquisition improvements on the resolution of EO products.
The PROBA-V satellite was launched on the 6th of May 2013 and is designed to provide space-borne vegetation measurements filling the gap between SPOT-VGT (from March 1998 to May 2014) and Sentinel-3 (April 2018).
The orbit of PROBA-V is at the altitude of \SI{820}{km} with a sun-synchronous inclination.
Interestingly, the payload includes sensors that ensure a near-global coverage (90\%) acquired with a \SI{300}{m} nominal resolution daily. The central camera observing at \SI{100}{m} nominal resolution, needs around 5 days to acquire a complete global coverage.
Consequently, images from the same land patch are available both at \SI{300}{m} resolution and, less frequently, at \SI{100}{m}.
This creates an ideal setup for a supervised machine learning scenario, in which multiple \SI{300}{m} resolution images can be labeled by a \SI{100}{m} resolution image relatively close in time.
The temporal differences between the images introduce complications like varying cloud coverage and minor changes in the scene, which require post-processing, but can in principle be utilized for the training of machine learning models. Thus, our objective is to explore the potential and limitations of this dataset for Super-resolution.

The contributions of this work are as follows: we describe the post-hoc assembly of a dataset from PROBA-V products containing different spatial resolutions, including information about cloud coverage. As a proof of concept, we train a simple multi-layer convolutional neural network to generate a super-resolved image out of a series of low resolution images and analyze its quality. We find the dataset to be rich enough to provide a scientific challenge for diverse approaches to Super-resolution, which is why we release our dataset to the public in form of a competition on the ESA's Advanced Concepts Team Kelvins\footnote{\url{http://kelvins.esa.int}} portal.

Our work is structured as follows: in Section~\ref{sec:related} we describe relevant recent work on SR in general and for EO in particular. We then describe, in Section~\ref{sec:data_collection}, the assembly of the image dataset and in Section~\ref{sec:quality} the metric we decided to use to assess the reconstruction quality. In Section~\ref{sec:experiments}, we describe the CNN architecture, its training and compare its super-resolved images to a bicubic upscaling. We conclude and give discuss possible improvements in Section~\ref{sec:conclusions}.

\section{Related Work}
\label{sec:related}

\subsection{Single Image Super-resolution}
\label{subsec:SISR}

Most work on SISR is concerned with the reconstruction of an artificially degraded image, usually by downscaling, blurring or other operators that emulate image acquisition in cameras. A overview about early approaches to SISR is described in the work of Yang et al.~\cite{Yang2014Single, Yang2010Image}, that provides a comprehensive benchmark. Popular algorithmic approaches, which are  mentioned by Yang et al. include:

\begin{itemize}
\item \textbf{Interpolation-based Models} (non-uniform upscaling, i.e. bilinear, bicubic or Lanczos resampling)
\item \textbf{Edge-based Methods} (training on various edge features, gradient profiles, etc.)
\item \textbf{Image Statistical Methods} (exploitation of image properties like heavy-tailed gradient distributions and regularization)
\item \textbf{Example-based/Patch-based Methods} (training of a learning function between cropped patches of the image with weighted averages, kernel regression, support vector regression, Gaussian process regression, sparse dictionary representation and similar).
\end{itemize}

Recent advances in networks had a significant impact on SISR. A landmark paper is the work of Dong et al.~\cite{Dong2016Image}, which were among the first to introduce deep neural networks to the field by proposing the SRCNN network architecture, largely outperforming previous approaches. SRCNN consists of an adaptable number of convolutional layers and multiple filters of different sizes which perform patch extraction, non-linear mapping and reconstruction. The success of SRCNN triggered a surge in research and lead to a stream of improvements~\cite{Kim2016Accurate,Dong2016Accelerating,Shi2016Real} and alternative CNNs resulting in better image quality and faster computation, mitigating some of the shortcomings of the initial design.
Recent neural network models like Generative Adversarial Networks (GANs) have been applied successfully to SISR be Ledig et al.~\cite{Ledig2017Photo}, achieving an up-scaling by a factor of 4 resulting in photo-realistic quality. GANs can fabricate image details that look plausible for a human observer, but have never been part of the original image. This features is useful for tasks like image inpainting~\cite{yeh2017semantic}, but might not necessarily be desirable for SISR.

Since 2017, the NTIRE challenge~\cite{Timofte2018NTIRE} established itself as one of the major benchmarks for SISR. The challenge is based on the DIV2K dataset~\cite{Timofte2017Ntire}, a open dataset of hand-selected and diverse images obtained from the internet. Those images are artifically degraded (by adverse conditions emulating realistic camera models or simple bicubic downscaling) and delivered as training pairs for learning.
According to NTIRE, the current state of the art in SISR is dominated by deep learning. The most successful methods are based on network architectures like the ResNet~\cite{He2016Deep} or the DenseNet~\cite{Huang2017Densely} architecture, which proved advantages for neural network based image processing in general.

The NTIRE challenge is an excellent example for the effectiveness of competitions to advance the state-of-the art. However, the dataset of NTIRE is not concerned with earth observation data and the evaluation of quality is (partly) biased towards ``good looks'' by incorporating quality metrics like SSIM~\cite{Wang2004Image}, that have been inspired by human perception. As this is undesirable for the processing of remote sensing data, the creation of a different type of dataset and challenge seemed needed and motivated our current work.

\subsection{Multi image Super-resolution}
\label{subsec:MISR}

Work on MISR is sometimes referred to as ``Multi Frame Super Resolution" because of its close relation to video processing. A common task is to infer a single high resolution image from a video originating from a source of low fidelity~\cite{Schultz1996Extraction, Faramarzi2016Blind}.
This typically implies a form of motion model (of the camera or the scene) that needs to be estimated either implicitly~\cite{Takeda2009Super} or explicitly, i.e. by optical flow~\cite{Mitzel2009Video} to provide accurate pixel registration.

While SISR is a mathematically ill-posed problem, in which lost pixel-information is somewhat imputed to generate a plausible and pleasing visual outcome, challenges in MISR can have a different objective: provided the multiple images are not \emph{exactly} identical, subpixel differences may allow to obtain actual information on the ground truth image. This makes the approach particularly attractive for EO satellites as it allows to enhance their payload capabilities keeping the information content of the newly created pixel values linked to real observations.

In comparison to MISR for video enhancement, the situation for EO is slightly different, as often multi-spectral data at varying resolutions is readily available. Brodu~\cite{brodu16} presents a super-resolution method which exploits the local and geometric subpixel consistencies across multiple spectral bands to enhance spectral bands of Sentinel-2 images of lower resolution. In particular, high resolution data are used to unmix the low-resolution bands in order to apply a technique similar to panchromatic sharpening~\cite{gillespie1987color, Thomas2008Synthesis}. Similar works on multi-spectral resolution enhancements have been published for SPOT-5~\cite{latry2009staggered}, ZY-3 TLC~\cite{li2017super} and SkySat-1~\cite{murthy}. These approaches, to the best of our knowledge, have been limited so far to data originating from the same satellite pass within an instant point in time and rely on panchromatic images, or at least images from different spectral bands. In contrast, our work explores the possibility of a temporal super-resolution inside a single spectral band, i.e. the improvement of image resolution by fusing images from multiple successive revisits of the satellite. In this sense, our task is more related to the MISR and SISR works discussed before, and less to pan-sharpening or image synthesis techniques.

\section{Data Collection}
\label{sec:data_collection}

\subsection{PROBA-V Data}
\label{subsec:probavdata}

We collect the satellite data from the free section of the PROBA-V product distribution portal\footnote{\url{https://www.vito-eodata.be/PDF/portal/Application.html\#Home}} of the European Space Agency. We downloaded the data available at product level L2A, which is composed of radiometrically and geometrically corrected Top-Of-Atmosphere (TOA) reflectances in Plate Carr\'{e} projection. The PROBA-V data is available for multiple spectral bands and resolutions, which are thoroughly described in the product manual~\cite{wolters2014proba}. 

The mission of PROBA-V is to monitor vegetation on Earth. For this purpose, it is important to compute the Normalized Difference Vegetation Index (NDVI), a quantity frequently deployed in practice for measuring biomass, indicating droughts, leaf density in forests and similar~\cite{carlson1997relation, pettorelli2005using}. Computing the NDVI requires information from the RED and NIR spectral band, which is why select them as the source of our data. For a given location, the RED and NIR channels provide (almost) daily images in \SI{300}{m} resolution and images in \SI{100}{m} resolution on average every 5 to 6 days.

From the \SI{300}{m} resolution data, we extract patches of 128x128 grey scale pixel images and refer to them as low resolution (LR) in the following. The corresponding \SI{100}{m} resolution data provides an upscaling by a factor of 3. Thus, we are able to extract patches of 384x384 pixels, covering the same region but in high resolution (HR). Additionally, we download a status map containing information on the presence of clouds, cloud shadows and ice/snow covering. The generation of this status map follows multiple procedures, applying pixel classification by incorporating information from ESA's Land Cover Climate Change Initiative and ESA's GlobAlbedo surface reflectance data. We refer to Section 2.2.3. of the PROBA-V product manual~\cite{wolters2014proba} for further details. The purpose of the status map for us is to identify suitable pixels for super-resolution as clouded pixels would provide false information. In the following, we call pixels identified as clouds by the status map \emph{concealed} and all other pixels \emph{clear}. Furthermore, we define the fraction of all clear pixels of an image as its \emph{clearance}.

\subsection{Spatio-temporal Selection of Regions of Interest}
\label{subsec:spatiotemporal}

\begin{figure}
\centering
\includegraphics[width=0.90\columnwidth]{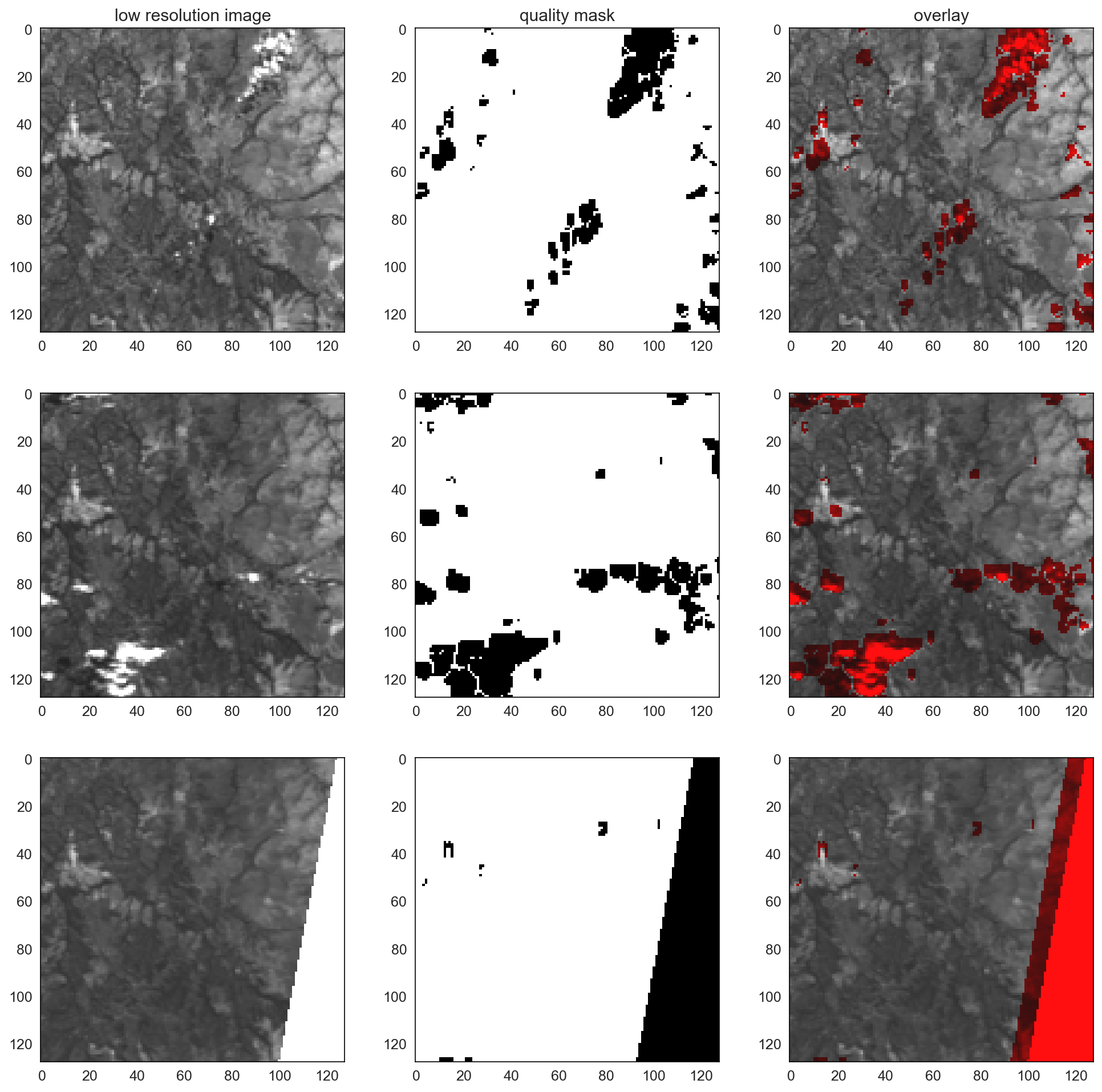}
\caption{The quality mask identifies concealed pixels, mostly related to clouds, cloud shadows and water vapor. The last row shows an image on the edge of the satellite swath. Missing pixels and the border-regions of the swath are identified as concealed.}
\label{fig:clvl}
\end{figure}

\begin{figure}
\centering
\includegraphics[width=0.98\columnwidth]{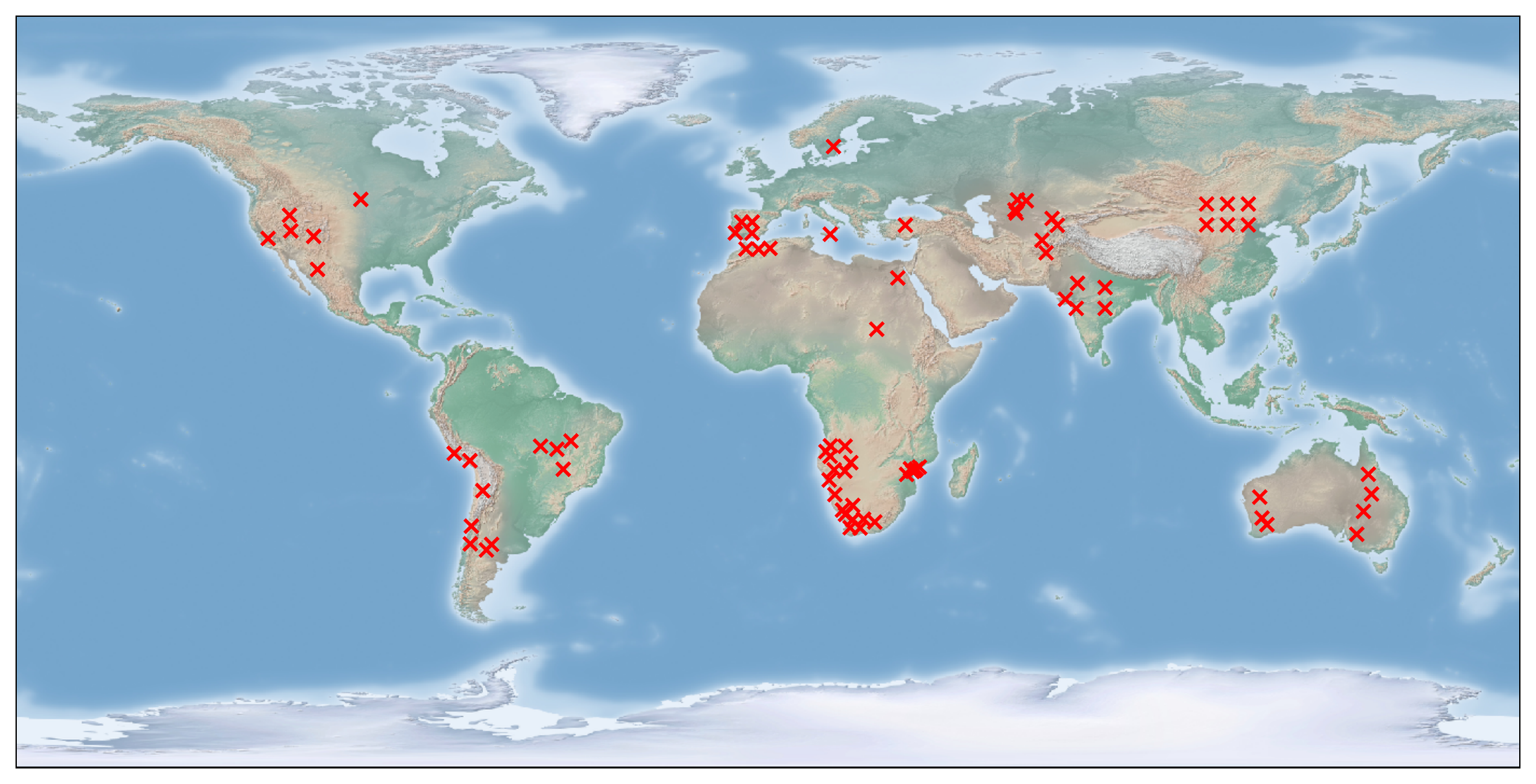}
\caption{Coordinates of the 74 regions, which were manually selected around the globe. Selection was guided to provide a diverse set of interesting vegetational and geological features with a low expected cloud coverage.}
\label{fig:rois}
\end{figure}

\begin{figure}
\centering
\includegraphics[width=0.98\columnwidth]{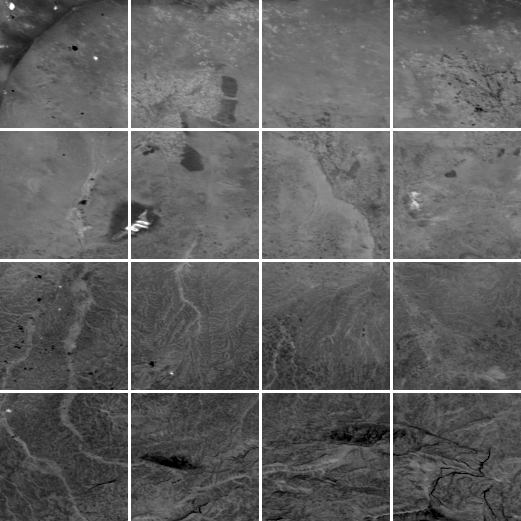}
\caption{Example of a 4x4 NIR image tiling of one ROI, located at 36$\degree$19'19.2"N 65$\degree$36'39.6"E, North Afghanistan.}
\label{fig:roiexample}
\end{figure}

For obvious reasons, images with a high clearance are better suited for super resolution, as any concealed pixel corresponds to a loss of information. Consequently, our selection for regions of interest (ROIs) should avoid areas that are frequently covered by clouds. We utilized a data-set collected by Wilson and Jetz~\cite{Wilson2016Remotely} about the annual global \SI{1}{km} cloud coverage to guide our manual selection. In total, we identified 74 different, land-based ROIs as shown in Figure~\ref{fig:rois}. Each ROI consists of 4x4 tiles in two spectral bands (RED and NIR), leading to 2\,368 (74x4x4x2) tiles. Given PROBA-V's resolution, one tile equates to roughly a land patch of \SI{1\,475}{km^2}. Figure~\ref{fig:roiexample} shows an example for a typical 4x4 tiling of a ROI.

We choose a time window of 30 days (one full month) for each ROI, for which we initially collect all available low and high resolution data. Ideally, one would choose this time window as small as possible, since a small time window makes changes in the landscape (for example due to swelling of rivers, harvests or seasonal changes in vegetation) less likely to occur. However, the availability of data varies and although we select the month for each ROI individually such that the expected cloud coverage is minimal, clouds still occur and may, in the worst case, conceal the whole tile during one day. Thus, the 30 day time window is a necessary trade-off between the quality and quantity of the images.

\subsection{Assembling a Dataset Guided by Clearance}

Given this time windows, we collect multiple high and low resolution images for each tile which we filter for favorable cloud coverage conditions. In particular, we discard any LR images with a clearance below $0.6$ (i.e. more than 40\% of the image is covered in clouds). For HR images, we are more strict and disregard everything below a clearance of $0.75$. The reasoning behind the stricter threshold is due to the fact that in our final dataset there will be only one HR image per datamember (to have a single target), but multiple LR images. Thus, a concealed region in one LR image might be clear in another image, while for a concealed pixel in an HR image no other source of truth is provided.

If a tile can not provide at least one HR image and at least nine LR images with the corresponding clearances, we do not include any information from the tile into our dataset. The values of $0.6$, $0.75$ and $9$ (for the number of LR images) have been picked to strike a favorable balance between quantity and quality of our data. Given those values, our selection procedure discards roughly 39\% of tiles, resulting in a final dataset of size 1\,450 (2\,368 - 918) tiles. Stricter requirements on the clearance would have reduced the number of datamembers significantly. This would have posed an issue for the application of CNNs, which are known to require comparatively large sets of images for training. We deemed 1\,450 members as sufficient suggested by the success of the NTIRE challenge, whose dataset consisted of 1\,000 images and thus selected our parameters accordingly.

Given the clearance of $0.75$ for HR images, PROBA-V's coverage interval of approx. $5$ days and the total time window of $30$ days, it happens that there is more than one valid HR image available for a tile. This is problematic, as for the intended purpose of our dataset (supervised learning), there should only be one HR target per datamember. Thus, in such situations we select the HR with the highest clearance. In case of a tie in clearance, we compute the pixelwise mean-square error of a downscaled (local-mean, zero-padding) version from each of those HR images with the set of available LR images, selecting the one that minimizes the error.

To summarize, out of 2\,368 tiles, we extracted a dataset with 1\,450 datamembers, where each member consists of the following set of images:

\begin{itemize}
\item a single high resolution target image, 384x384 pixels, 16 bit grey-scale
\item a corresponding high resolution quality mask, indicating concealed pixels (at least 0.75 clearance), 384x384 pixels, 1 bit.
\item multiple (at least nine) low resolution images, 128x128 pixels, 16 bit grey-scale.
\item low resolution quality masks for each low resolution image, indicating concealed pixels (at least 0.6 clearance), 128x128 pixels, 1 bit.
\end{itemize}

\section{Quality Metric}
\label{sec:quality}

Many of the most popular image quality metrics (like the structural similarity index (SSIM)~\cite{Wang2004Image} or the Visual Information Fidelity (VIF)~\cite{Sheikh2004Image}) have been inspired by human perception. This is understandable, as many super-resolution applications are concerned with delivering high resolution primarily for the human visual system for consumption. However, this human perception bias makes them less applicable to scientific EO tasks, where the visual representation might be less important than, for example, accurate pixel values. In order to select a suitable quality metric, we have to clarify the specific requirements that should be reflected by our evaluation.

\subsection{Ranking Super-resolved Images}
We note the following high-level requirements for a super-resolved (SR) image:

\begin{enumerate}
\item The pixel-wise values in SR should be as close as possible to the target HR image after the removal of unnecessary bias.
\item The quality of the image should be independent of pixel-values marked as concealed in the target image.
\item The SR image should not reconstruct volatile features (like clouds) or introduce artifacts.
\end{enumerate}

Given these requirements, we take the Peak Signal Noise Ratio (PSNR) as our starting point, but modify a few aspects to address certain properties of our data. 

For the test set we have to compare generated test images (SR) with the ground truth high resolution images (HR), both are 384x384px images. The geolocation mean accuracy of PROBA-V depends on the channel, but is always roughly around 60m ($\pm$50m standard deviation). Consequently, minor shifts in the content of the pixels are expected. One might argue that those random subpixel-shifts could be exploited for the task at hand which is why we refrain from fine image registration at the level of the data. Instead, we design our quality metric to have some tolerance for small pixel-translations in the high-resolution space by evaluating on a sliding cropped image (i.e. we look for a displacement of SR by at most 3 pixels in each direction that minimizes the error).

Formally, we denote a 378x378px cropped image as follows: for all $u, v \in \{ 0, \ldots, 6 \}$, $\text{HR}_{u,v}$ is the subimage of HR with its upper left corner at coordinates $(u,v)$ and its lower right corner at $(378 + u, 378 + v)$. Analogously, $\text{SR}_{3,3}$ is the center part of SR which is acquired by cropping a 3px border. In the following, we will omit the $(3,3)$ offset in the notation of SR. The pixel-intensities of those images are represented as real numbers $\in [0,1]$ for any given image $I$, obtained by a linear scaling from the (raw) integer values of the original 16bit representation. Furthermore, let $clear(I)$ be the set of pixel coordinates that are indicated as clear for image $I$ by its corresponding quality map.

\begin{figure}
	\centering
	\includegraphics[width=0.98\columnwidth]{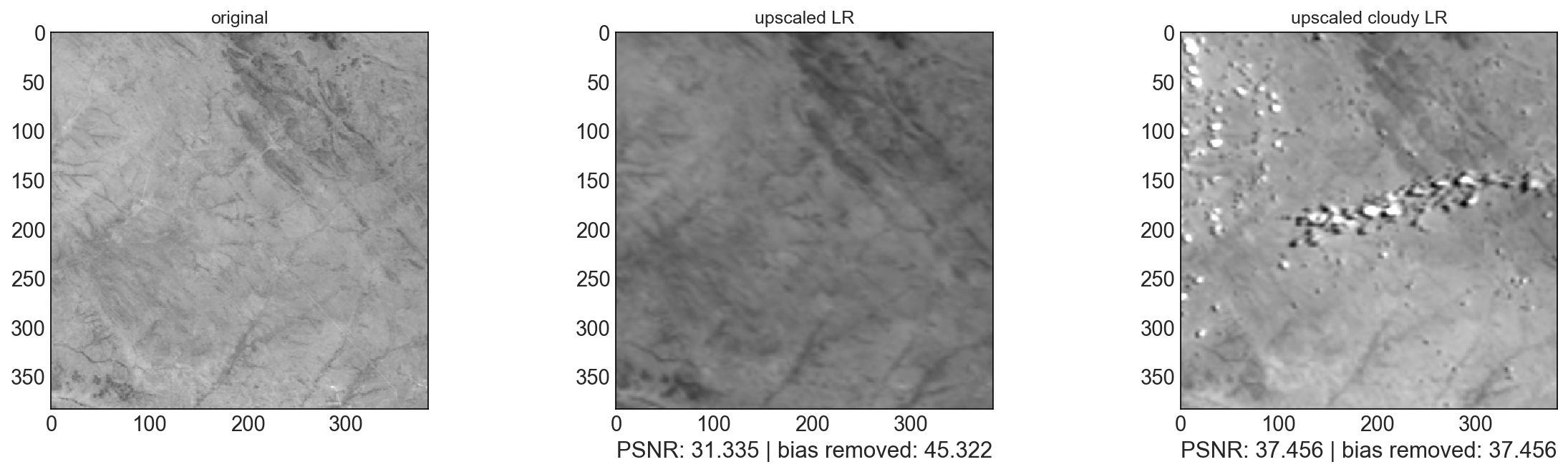}
	\caption{From left to right: original HR image, clear LR image and clouded LR image. The average pixel intensities for HR and clouded LR are roughly the same with 0.1386, while the clear LR has a lower average pixel intensity with 0.1120. Without correcting for this bias, the PSNR ranks the clouded image significantly higher than the clear image.}
	\label{fig:psnrweak}
\end{figure}

We note, that if SR = HR + $\varepsilon$ (i.e. each pixel-intensity of SR equals the corresponding pixel intensity of HR plus a constant bias) a perfect reconstruction of HR is possible if $\varepsilon$ is known and no over or under-saturation occurs. Thus, a quality metric reflecting the above requirements should award SR a high score in comparison to reconstructions that, for example, introduce noise to HR and thus irrevocably erase information. As it turns out, there are conditions under which the PSNR does the exact opposite and punishes a bias in intensity much more severe than the occurrence of noise.

Figure~\ref{fig:psnrweak} shows an example of two upscaled LR images, one with features not present in the HR ground truth (clouds) but with the same average pixel intensity and one with perfect clearance, but a bias in intensity. Note, that there are no clouds in the HR image, which means that each pixel of the HR image is clear and thus needs to be super-resolved. Since we can compensate for a constant bias in intensity much more easily than for a reconstruction containing noise or troublesome features like clouds, we soften our quality metric by removing bias from SR before computing anything else.

More precisely, for every possible $u,v$, we compute the bias in brightness $b_{u,v}$ as follows:

\begin{equation}
    b_{u,v} = \frac{1}{\mid clear(\text{HR}_{u,v})\mid} \sum\limits_{\{x, y\} \in clear(\text{HR}_{u,v})} \left( \text{HR}_{u,v}(x,y) - \text{SR}(x,y) \right).
\end{equation}

This correction ranks the example given by Figure~\ref{fig:psnrweak} in better correspondence to our requirements. For similar reasons as mentioned before, we refrain from correcting this bias at the level of the data and instead decrease the sensitivity of PSNR in our quality metric by incorporating the bias term in the computation of the mean square error. Additionally, we exclude all concealed pixels of HR from the error, leading to the corrected clear mean-square error MSE of SR w.r.t. $\text{HR}_{u,v}$ given by:

\begin{equation}
    \text{MSE}(\text{HR}_{u,v}, \text{SR}) = \frac{1}{\mid clear(\text{HR}_{u,v})\mid} \sum\limits_{\{x, y\} \in clear(\text{HR}_{u,v})} \left(\text{HR}_{u,v}(x,y) - (\text{SR}(x,y) + b_{u,v}) \right)^2
\end{equation}

Given this modified MSE, we compute the PSNR for a particular displacement $u,v$ as

\begin{equation}
    \text{PSNR}(\text{HR}_{u,v}, \text{SR}) = -10 \cdot \log_{10} \left(\text{MSE}(\text{HR}_{u,v}, \text{SR})\right).    
\end{equation}

Applying image registration to the PSNR gives our final score, which we call cPSNR:

\begin{equation}
    \text{cPSNR}(\text{HR}, \text{SR}) = \max\limits_{u,v \in \lbrace 0, \ldots, 6 \rbrace} \left( \text{PSNR}(\text{HR}_{u,v}, \text{SR}) \right).
\end{equation}

\subsection{The Impact of Temporal Differences Between Images}
\label{sec:change_scene}

\begin{figure}
	\centering
	\includegraphics[width=0.98\columnwidth]{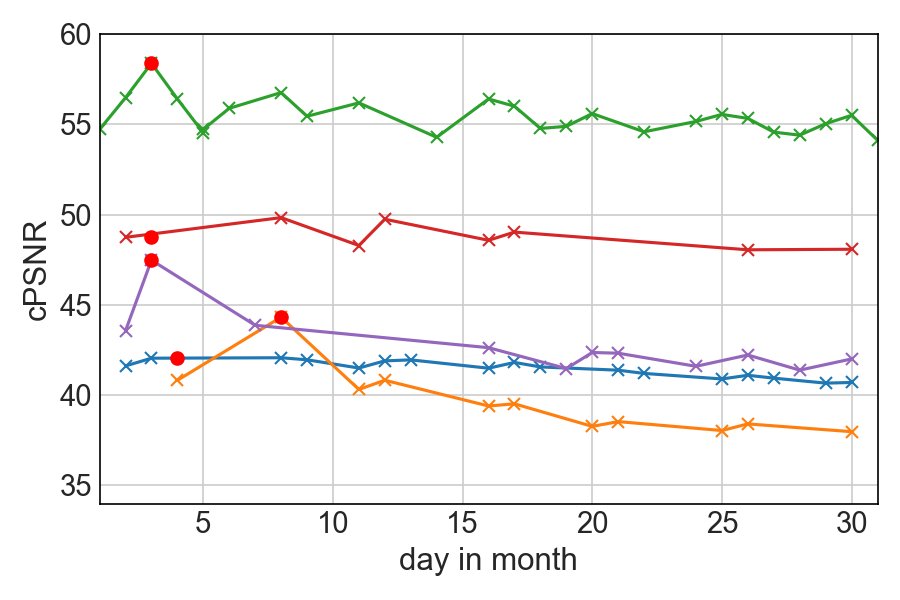}
	\caption{Multiple images from five different scenes, filtered for complete clearance and corrected for illumination bias. Each colored line is a different scene. The red circles mark the day of the reference HR image, while each cross is a cPSNR between this HR reference and an upscaled LR image recorded on the corresponding day.}
\label{fig:changing_landscape}
\end{figure}

Differences in the observed scene (e.g. due to weather effects, human interventions or changing seasons) are to be expected in an EO data-set collected over the time-scale of a month. The question arises, how these temporal differences will impact the cPSNR?

While a more careful selection of ROIs or a shorter time window could mitigate the issue somewhat, there is little escape from it in the scenario of multiple revisits.
Our investigations show that the change in cPSNR with each successive satellite revisit depends highly on the specifics of the scene and the temporal distance of its images.
Thus, it is difficult to draw conclusions without taking the complex particularities of the individual scenes into account. A matter which is further complicated by the fact that any decrease in cPSNR could also be attributed to other factors (such as registration, illumination bias or clouds) and the ability of any candidate super-resolution algorithm handle each of those.

To study the impact of temporal changes in isolation, we had to analyze the few images that had an almost ($>$99\%) perfect clearance for the duration of the whole month. Figure~\ref{fig:changing_landscape} shows five examples of such images. For these data-members we observed a high variation in the temporal deviations of the cPSNR, ranging from $0$ up to $6$, depending on the content of the scene. While we cannot generalize from these examples to the whole dataset, it suggests that the temporal changes will have an impact on our quality metric. Further research is warranted, but would require a larger study of data,
beginning with a classification and quantification of temporal change which is not in the scope of this work. Judging from the examples presented in Figure~\ref{fig:changing_landscape}, it is entirely possible that an algorithm provided with temporal information (i.e. the exact time at which each HR and LR image were recorded) could increase in cPSNR by avoiding the error potentially introduced by larger temporal distances. However, our focus for the upcoming supervised learning is not to incorporate temporal or any other types of meta-information into the learning process, but to analyze purely what can be learned from pixels alone.

\section{Experiments}
\label{sec:experiments}

To demonstrate the suitability of the collected data for supervised machine learning, we implement a convolutional neural network similar in structure to the one proposed by Dong et al.~\cite{Dong2016Accelerating}, consisting of multiple convolutional layers operating in LR space and finishing with a deconvolution-layer for upscaling to HR space. As the original network of~\cite{Dong2016Accelerating} was designed for SISR, we modified the input layer to receive multiple images as different channels (MISR). 

After manually removing 7 data-members of poor quality (clearest image barely above the threshold), we split our data-set randomly into subsets of 290 members for testing and 1\,153 members for training. For the split we took care that no pair of corresponding NIR and RED images (images obtained from the same tile but from different spectral bands) were separated between the training and test set, to keep them independent. 

In the following, we refer to a \emph{datamember} as a set of images from our dataset consisting of multiple images and different image types: one HR target image, several LR images and the corresponding quality maps. Recall, that each datamember thus contains a snapshot of a landscape tile in either the RED or NIR spectral band sampled over the period of one month.

\subsection{Network Design Details and Training}
\label{subsec:training}

\begin{figure}
	\centering
	\includegraphics[width=0.98\columnwidth]{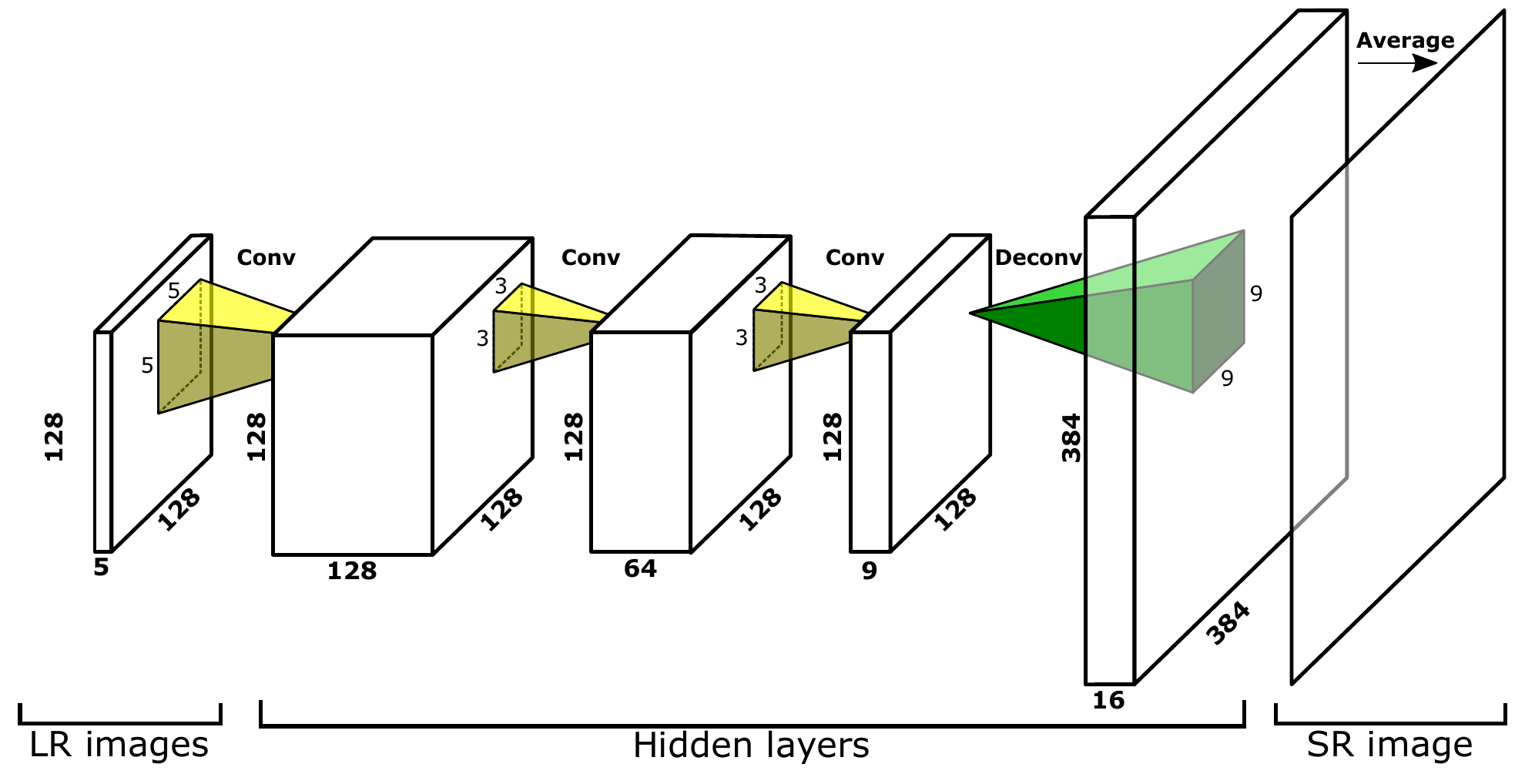}
	\caption{Illustration of the multi image super resolution (MISR) network architecture.}
	\label{fig:topology}
\end{figure}

The guaranteed number of LR images per datamember is (by the design of the dataset) nine. Most of the time, more images (on average 19) are available, but extending neural networks to deal with variably sized inputs is not straightforward. Additionally, many of the LR images have a low clearance which showed to have a detrimental impact on the training. Thus, after evaluating different alternatives, we settled for an input layer expecting exactly five 128x128 images, which have been selected for maximum clearance out of all available LR images per datamember. Each of those five images is handled as a different channel and is passed through multiple convolutional layers: The first layer transforms the images to $128$ feature maps using a 5x5 kernel, the second maps to $64$ feature maps using a 3x3 kernel, and the third layer maps to $9$ feature maps using again a 3x3 kernel. Each convolutional layer uses a stride of $1$ and padding to keep the image dimension at 128x128. The last layer of the network is a deconvolution layer with a fractional stride of $1/3$ that maps into $16$ channels in the 384x384 resolution space. The final super-resolved image is the average out of those $16$ channels. The activation function is \emph{ReLu} for all layers. Figure~\ref{fig:topology} shows an illustration of the network architecture. The total number of trainable biases and weights of the network is 119\,610.

The network is trained using the Adam optimizer with default parameters to minimize the mean square error over $200$ epochs. The learning rate is initialized with $0.001$ and decays exponentially down to $7.666\cdot10^{-5}$ during training. Preliminary experiments showed that our network learns faster with mini-batches of size $4$ on GPUs. 

\subsection{Results}
\label{subsec:results}

\begin{figure}
\centering
\includegraphics[width=0.49\columnwidth]{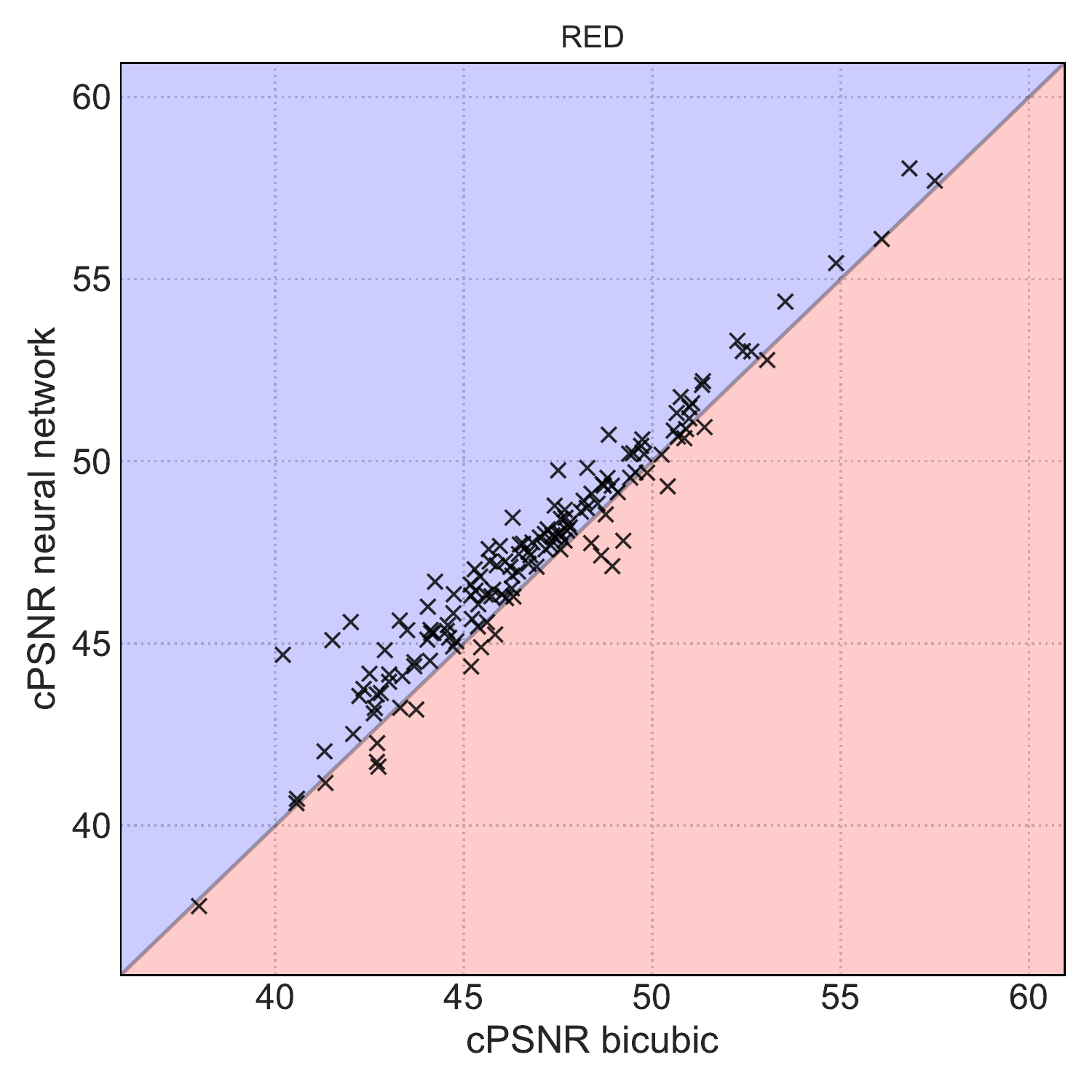}
\includegraphics[width=0.49\columnwidth]{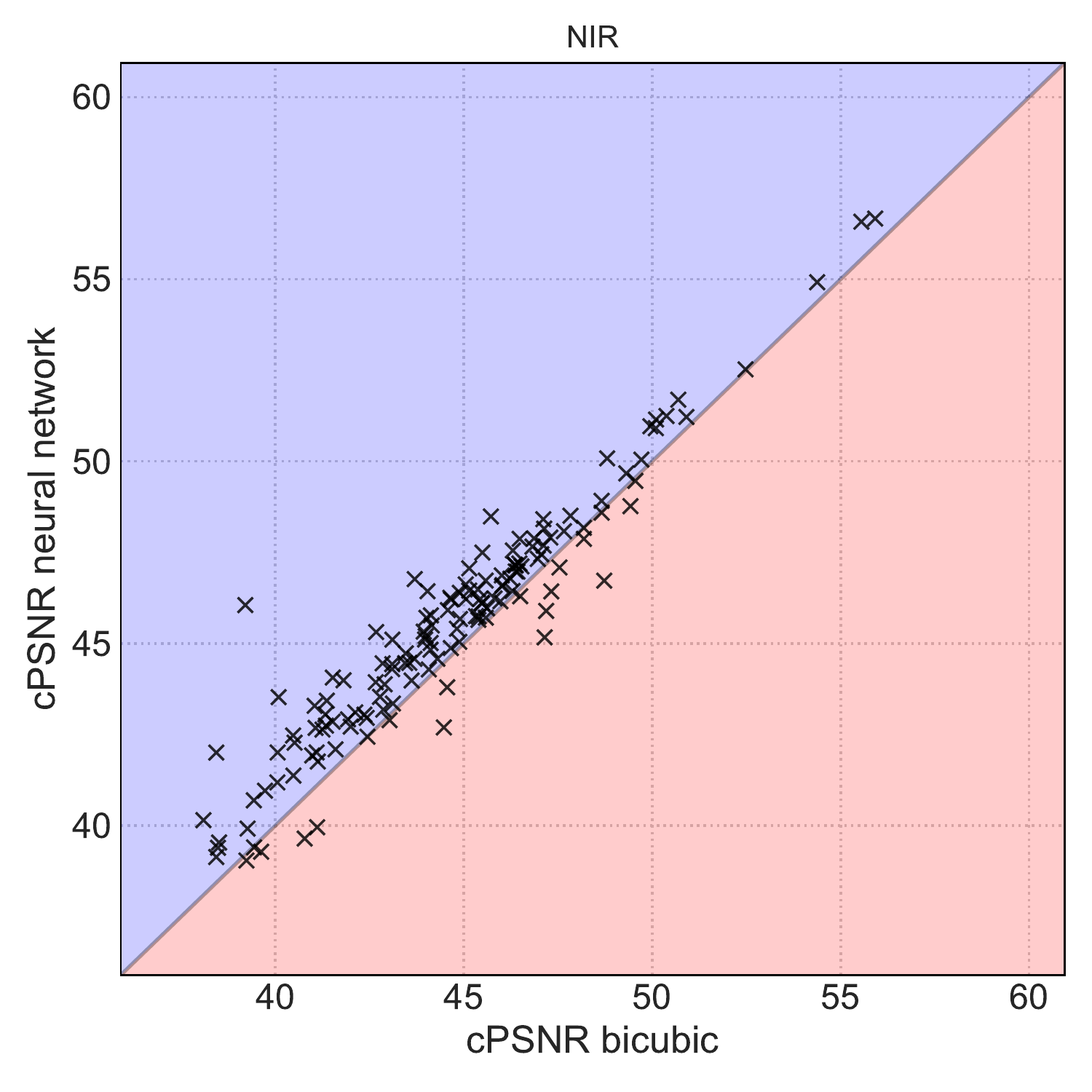}
\caption{Each cross represents an image of the test set. Crosses in the blue area mean that the neural network achieved a higher cPSNR than the average score of bicubic interpolations of the best available low resolution images.}
\label{fig:comp}
\end{figure}

To establish a baseline, we compute the average cPSNR for each member of the dataset by a bicubic upscaling of all LR images that have maximum clearance, assuming that these set of images have to be considered ``best choice'' in the absence of any further information. We compare this average cPSNR with the cPSNR that the neural network achieves. Figure~\ref{fig:comp} shows this comparison over the whole dataset divided into RED and NIR images. We note that our neural network achieves better cPSNR than the bicubic interpolation in 251 out of 290 test cases. Table~\ref{tab:avgcpsnr} breaks our results down for the RED and NIR images of our test-set and reports the average cPSNR.

\begin{table}
\centering
\def\arraystretch{1.2}
\setlength{\tabcolsep}{0.5em} 
\begin{tabular}{l|c|c|c}
spectral band & RED & NIR & RED + NIR \\ 
\hline 
avg. cPSNR bicubic & 46.782 & 44.659 & 45.728 \\ 
avg. cPSNR MISR network & 47.445 & 45.520 & 46.489 \\ 
total number of images & 146 & 144 & 290 \\ 
network better than bicubic & 121 & 124 & 251 \\ 
\end{tabular} 
\caption{Average performance of the MISR network compared to bicubic interpolation.}
\label{tab:avgcpsnr}
\end{table}

\begin{figure}
\centering
\includegraphics[width=0.98\columnwidth]{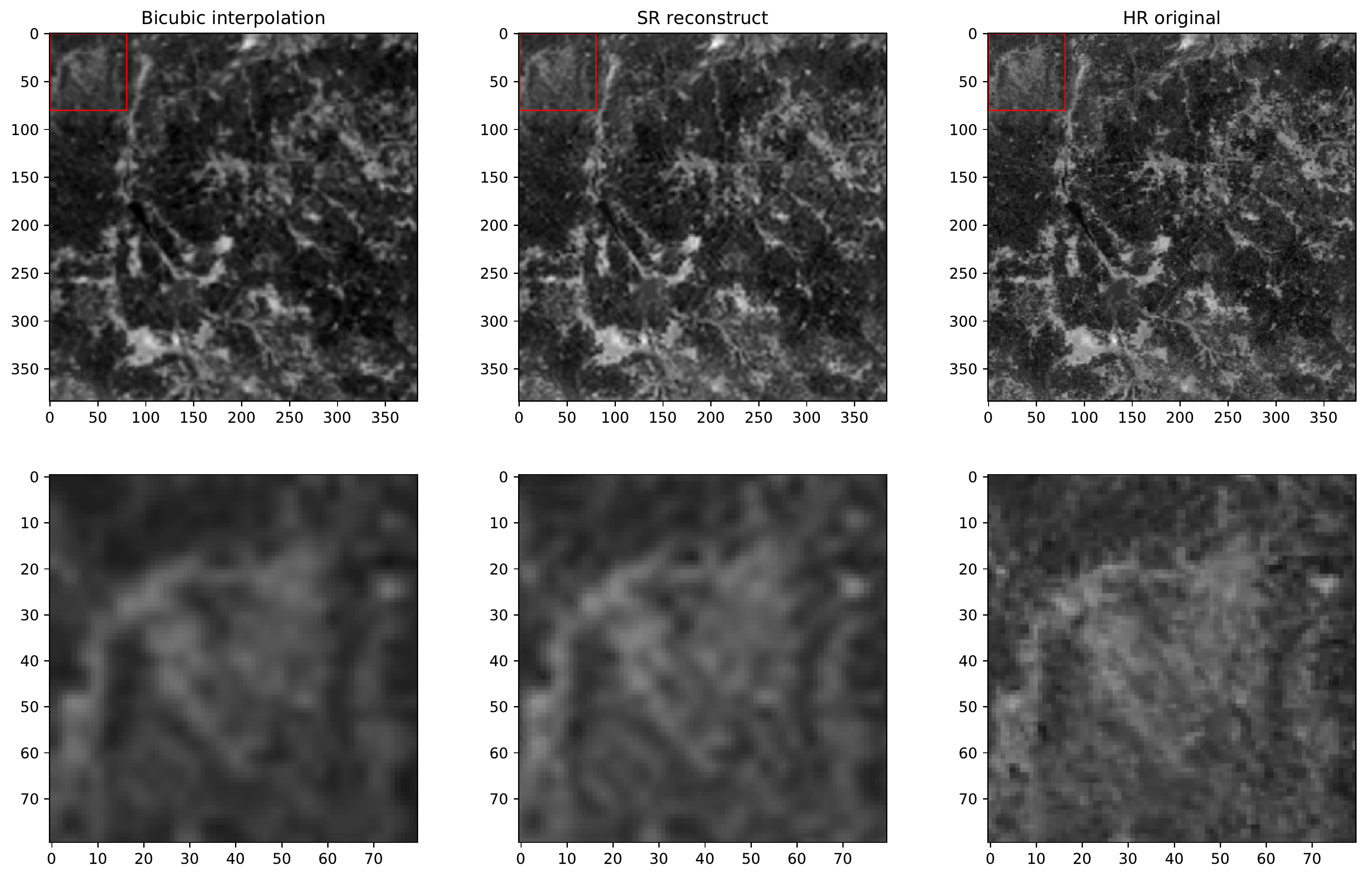}
\caption{cPSNR of interpolation: 45.6369. cPSNR of super-resolved image: 45.8022.}
\label{fig:samp53}
\end{figure}

\begin{figure}
\centering
\includegraphics[width=0.98\columnwidth]{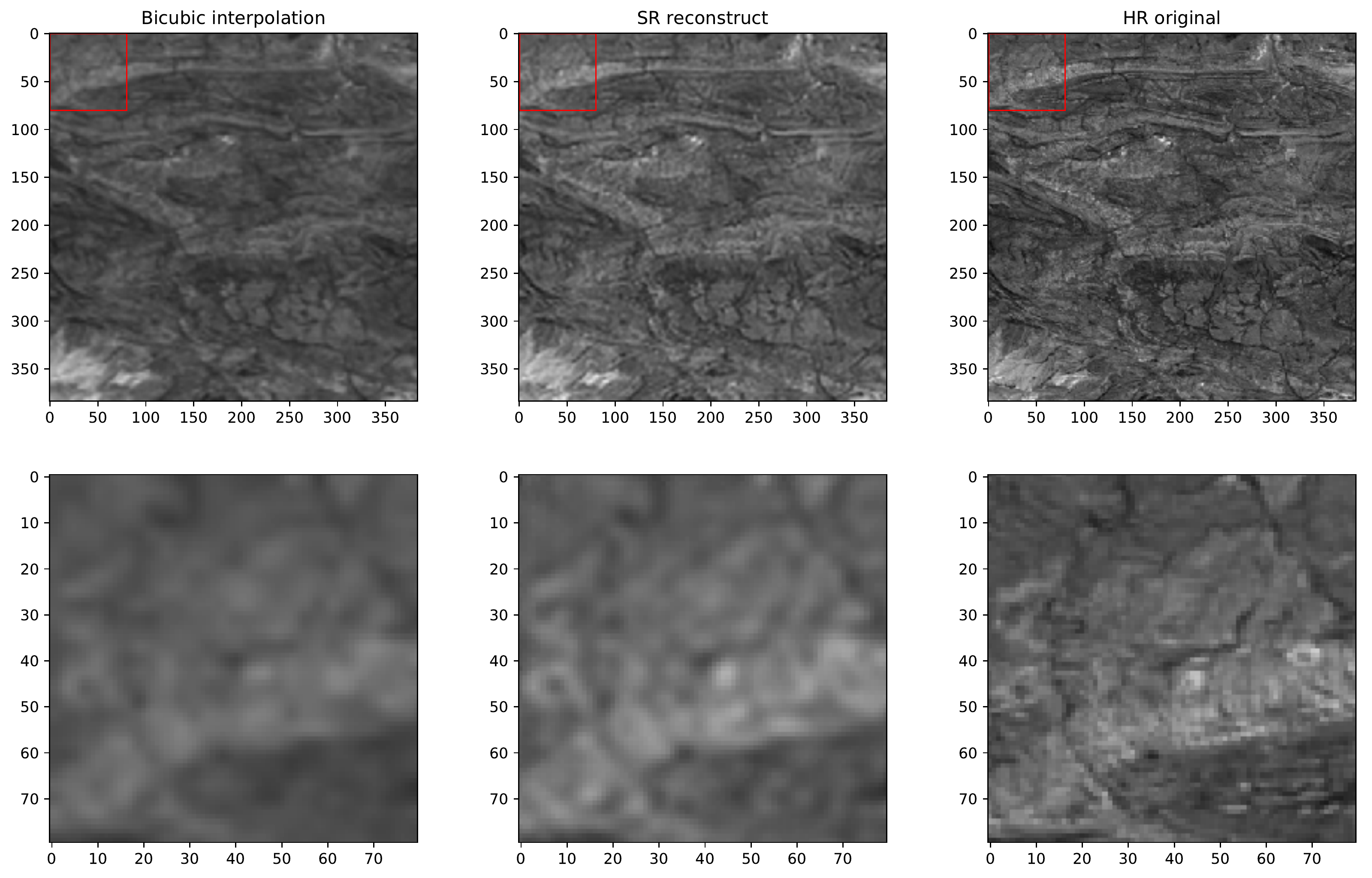}
\caption{cPSNR of interpolation: 45.5862. cPSNR of super-resolved image: 47.0913.}
\label{fig:samp78}
\end{figure}

\begin{figure}
\centering
\includegraphics[width=0.98\columnwidth]{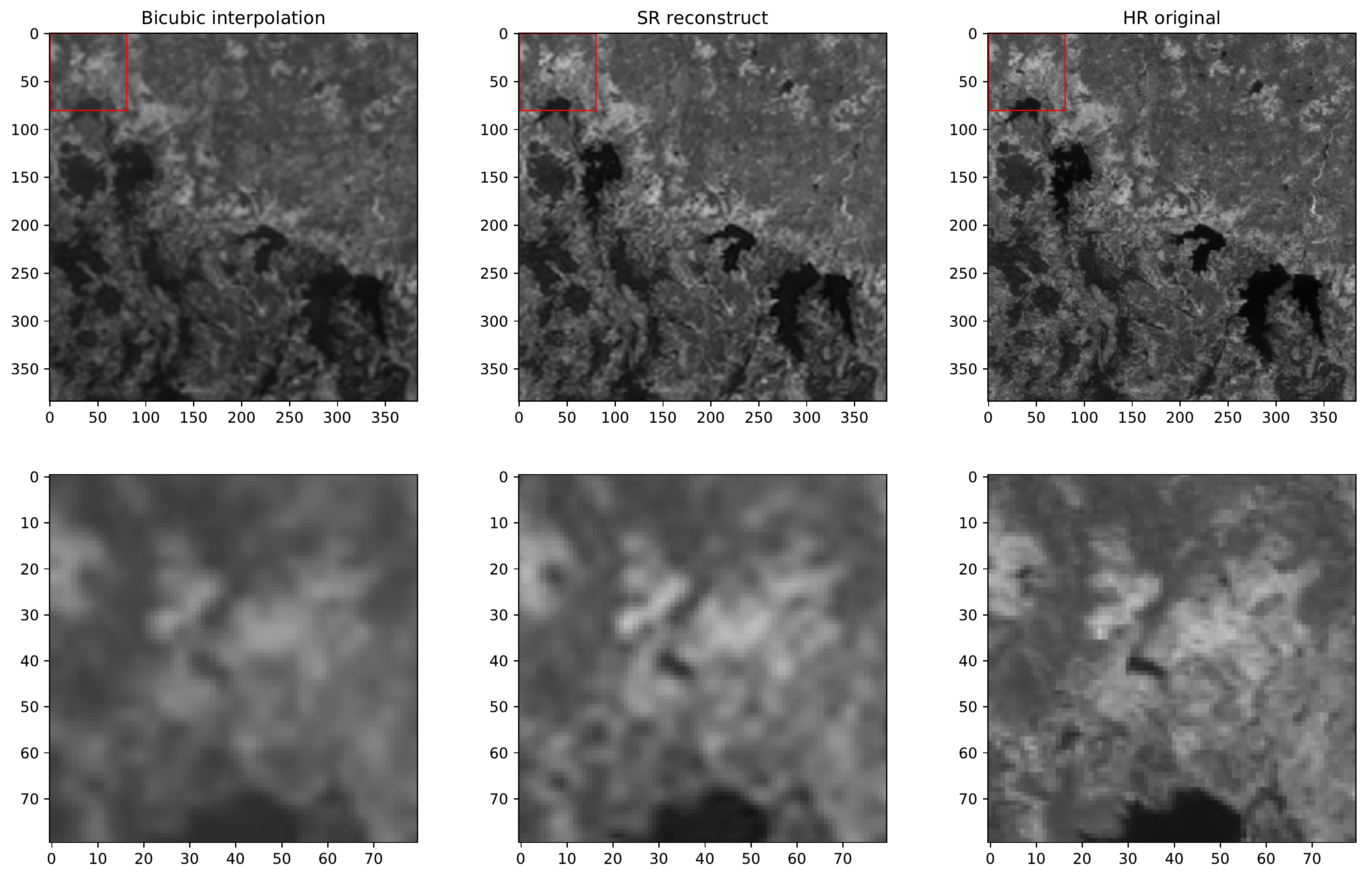}
\caption{cPSNR of interpolation: 46.2723. cPSNR of super-resolved image: 48.4106.}
\label{fig:samp113}
\end{figure}

\begin{figure}
\centering
\includegraphics[width=0.98\columnwidth]{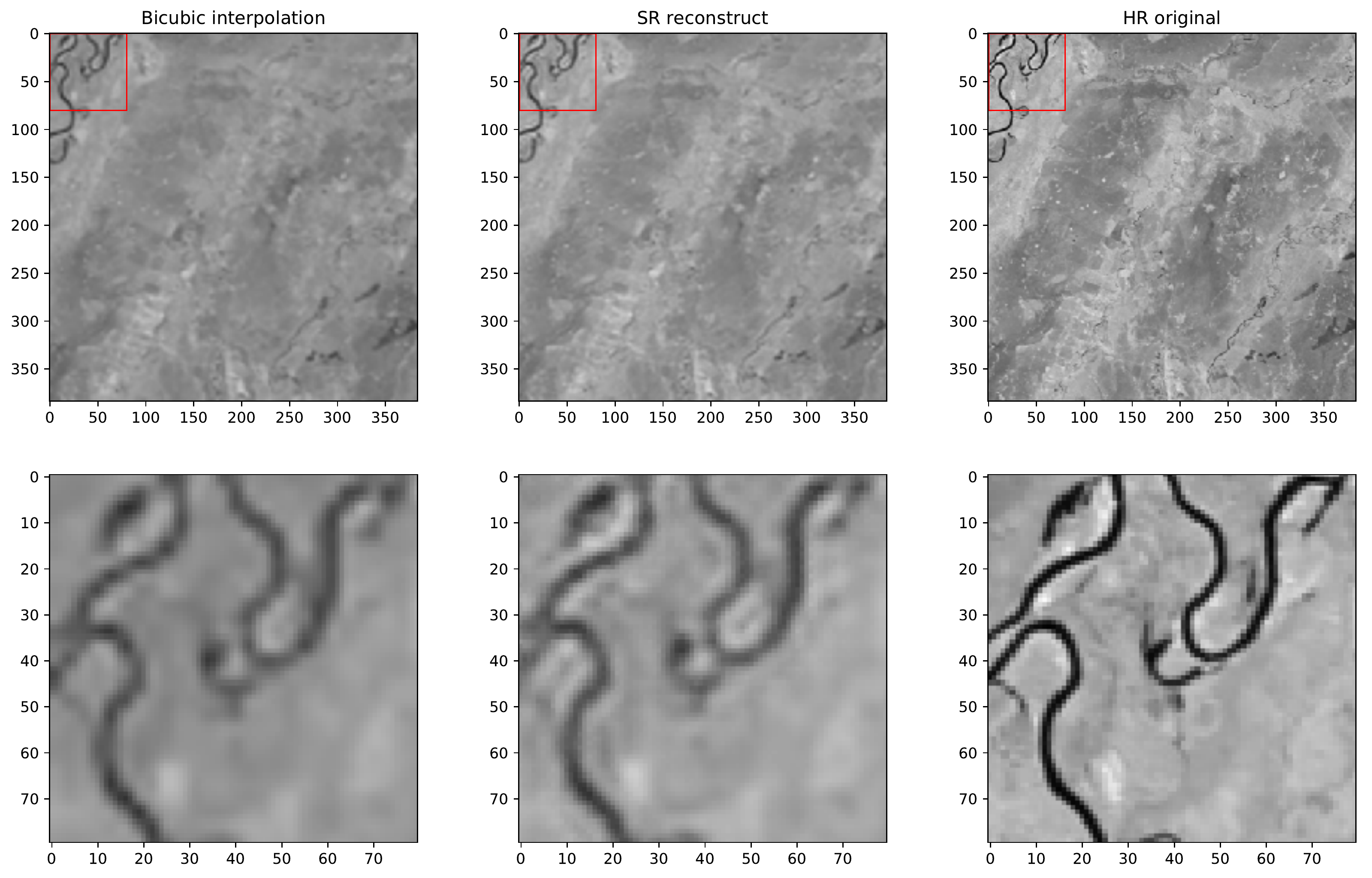}
\caption{cPSNR of interpolation: 40.5918. cPSNR of super-resolved image: 41.7992.}
\label{fig:samp224}
\end{figure}

\begin{figure}
\centering
\includegraphics[width=0.98\columnwidth]{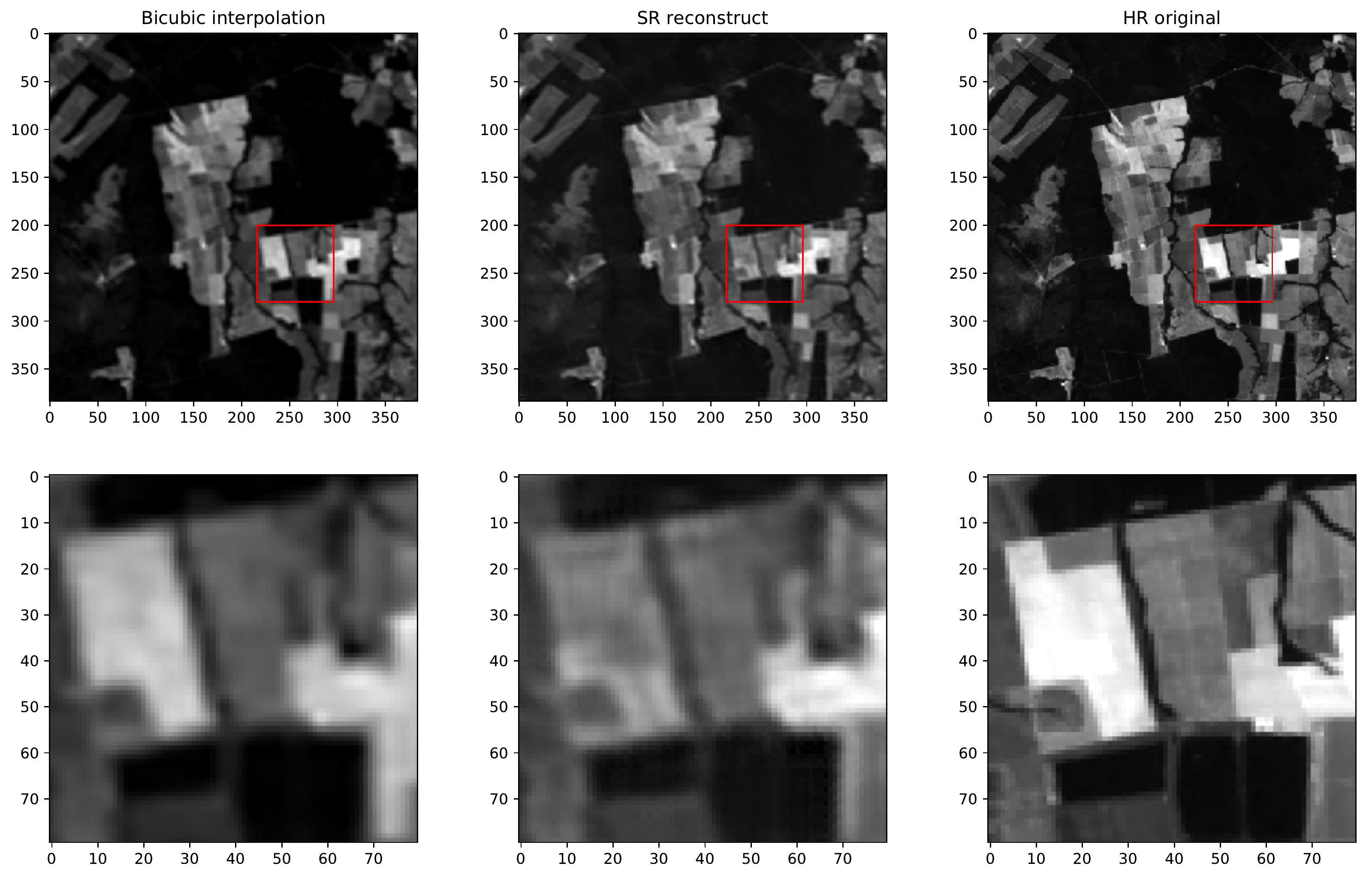}
\caption{An example where we focused the zoom-in on an area that changes over time. The differences between the LR images of this scene are comparatively large. While both, the SR reconstruct and the Bicubic interpolation clearly do not match regions of the HR target, the SR performs worse. cPSNR of interpolation: 46.7764. cPSNR of super-resolved image: 45.9042.}
\label{fig:counterex}
\end{figure}

Inspecting the results, we find that the SR reconstruction, in general, provides a visually higher quality than the bicubic interpolation.~\Cref{fig:samp53,fig:samp78,fig:samp113,fig:samp224} show some examples in comparison, with reconstructed images in detail and the corresponding cPSNRs. While the finer details of the HR target are not fully recovered, some higher level features like the edges of rivers become more visible when images are processed with the MISR network (compare Figure~\ref{fig:samp224}). Inspecting the few cases in which the MISR network provided a worse cPSNR than the bicubic interpolation often reveals a data-member with a comparatively higher amounts of temporal changes within the scene, which might be difficult for the network to handle. Figure~\ref{fig:counterex} shows one of these examples.

\section{Discussions and Conclusions}
\label{sec:conclusions}
Given the results presented in the previous sections, we conclude that our dataset - as provided - is already suitable for the training of deep convolutional networks. While Super-resolution has been applied to similar single pass scenarios before, our study delivers evidence that convolutional neural networks are powerful enough to achieve improvements in visual quality for multi-pass scenarios as well, potentially opening the door for post-acquisition enhancements for large amounts of EO data. With the cPSNR we introduced a new quality metric that is applicable for images which only have partial information (due to cloud coverage) and is less sensitive to pixel-shifts and constant biases in intensity opposed to the unmodified PSNR.

The MISR network which we proposed, while simple in design, provided already an improvement in cPSNR with respect to bicubic interpolation, which we see confirmed by a visual inspection of the results. We see our network as a first stepping stone, upon which multiple improvements can be developed and tested. For example, the network structure was not specifically optimized for the task at hand and could most certainly be improved by a hyperparameter optimization (number of layers, size of filters, learning rate schedules, weight initialization etc.). We deliberately limited the input of the MISR network to the 5 clearest images available in our dataset to mitigate the impact that clouds would have on the reconstruction process. Apart from that, our approach ignores the status maps, although we find that they provide a reliable source of information on the quality of the low resolution pixels. Furthermore, data augmentation techniques have not been investigated, but are straightforward to apply, for example by rotations, mirroring or cropping.

Lastly, as shown in Section~\ref{sec:change_scene}, the temporal dimension has an impact on the quality of the reconstruction, due to the accumulation of changes in the scene. However, our approach is agnostic when it comes to time: the selection of images is only guided by clearance. Incorporating the information on clear pixels directly into the network while exploiting temporal proximity could be another promising direction for further advancement. Looking at the few cases where the super-resolved images were of lower cPSNR than the bicubic interpolations almost always revealed data members that have undergone considerable scenic changes between the successive satellite passes.

To summarize, we conclude that multi-image super-resolution of satellite data over a multiple day time window is feasible and improves image quality over a bicubic interpolation baseline. While this post-hoc improvement has been proposed mainly with MISR in mind, it is possible that the convolutional neural networks developed for MISR could also be beneficial to correct images taken by satellites that have been disturbed by microvibrations. Ultimately, we think that the full potential of our data has not been completely exhausted. Thus, we made our dataset available to the public via the European Space Agency's competition platform Kelvins\footnote{\url{http://kelvins.esa.int}} with the goal to inspire further developments in machine learning or image processing algorithms.

\begin{acknowledgement}
The authors would like to thank the anonymous reviewers for their extremely detailed comments and critical questions, which helped to make the presentation of these results clearer and increased the overall quality of this work.
\end{acknowledgement}

\bibliographystyle{astrobib}
\bibliography{main}

\section{Authors short bios}
\label{sec:AUTHORS}

\subsection{Marcus M\"artens}

\begin{minipage}{0.6\textwidth}
Marcus M\"artens graduated from the University of Paderborn (Germany) with a Masters degree in computer science. He joined the European Space Agency as a Young Graduate Trainee in artificial intelligence where he worked on multi-objective optimization of spacecraft trajectories. He was part of the winning team of the 8th edition of the Global Trajectory Optimization Competition (GTOC) and received a HUMIES gold medal for developing algorithms achieving human competitive results in trajectory design.
\end{minipage}
\begin{minipage}{0.4\textwidth}
\centering
\vspace{0.2cm}
\includegraphics[width=0.9\textwidth]{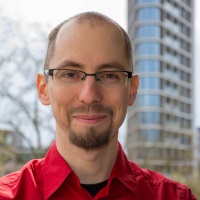}
\vspace{0.6cm}
\end{minipage}

The Delft University of Technology awarded him a Ph.D. for his thesis on information propagation in complex networks. After his time at the network architectures and services group in Delft (Netherlands), Marcus rejoined the European Space Agency, where he works as a research follow in the Advanced Concepts Team. While his main focus is on applied artificial intelligence and evolutionary optimization, Marcus has worked together with experts from different fields and authored works related to neuroscience, cyber-security and gaming. E-Mail: marcus.maertens@esa.int.

\subsection{Dario Izzo}

\includegraphics[width=0.9\columnwidth]{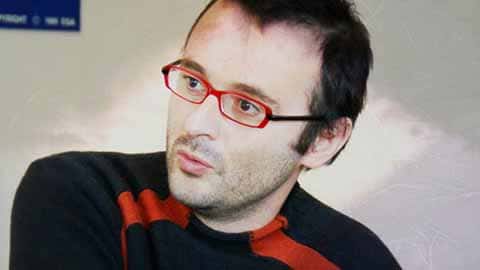}

Dario Izzo graduated as a Doctor of Aeronautical Engineering from the University Sapienza of Rome (Italy). He then took his second master in “Satellite Platforms” at the University of Cranfield in the United Kingdom and completed his Ph.D. in Mathematical Modelling at the University Sapienza of Rome where he lectured classical mechanics and space flight mechanics.

Dario Izzo later joined the European Space Agency and became the scientific coordinator of its Advanced Concepts Team. He devised and managed the Global Trajectory Optimization Competitions events, the ESA’s Summer of Code in Space and the Kelvins innovation and competition platform for space problems. He published more than 170 papers in international journals and conferences making key contributions to the understanding of flight mechanics and spacecraft control and pioneering techniques based on evolutionary and machine learning approaches.

Dario Izzo received the HUMIES Gold Medal and led the team winning the 8th edition of the Global Trajectory Optimization Competition. E-Mail: dario.izzo@esa.int.

\subsection{Andrej Krzic}

\begin{minipage}{0.6\textwidth}
Andrej Krzic graduated in physics from the University of Ljubljana (Slovenia) and afterwards specialised in optics by earning a M.Sc. degree from the Friedrich Schiller University Jena (Germany). While in Jena, he was working at Carl Zeiss Microscopy GmbH, where he was investigating applications of adaptive optics for microscopic imaging. After his studies, Andrej spent two years as a Young Graduate Trainee at the European Space Agency (ESA) in the Netherlands. \end{minipage}
\begin{minipage}{0.4\textwidth}
\centering
\vspace{0.2cm}
\includegraphics[width=0.8\columnwidth]{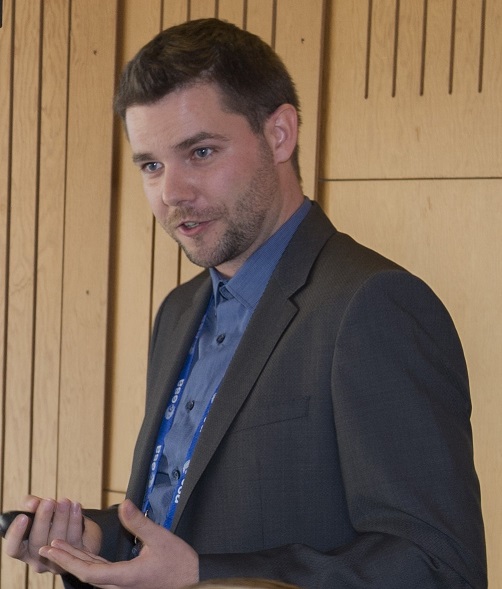}
\vspace{0.6cm}
\end{minipage}
He was part of the Advanced Concepts Team, where he was active in several areas of research, including super-resolution imaging, quantum metrology, and optical communication. Andrej later returned to Jena, where he is now working on adaptive optics assisted quantum communication at the Fraunhofer Institute for Applied Optics and Precision Engineering. E-Mail: Andrej.Krzic@iof.fraunhofer.de.

\subsection{Dani\"el Cox}
\begin{minipage}{0.6\textwidth}
Dani\"el Cox has graduated as a Bachelor of Science in Applied Physics at the University of Twente (The Netherlands) and is currently working on his Masters degree Applied Physics with a focus on optics and image processing. He has briefly been a part of the Advanced Concepts Team at the European Space Agency during his internship. E-Mail: d.w.s.cox@alumnus.utwente.nl.
\end{minipage}
\begin{minipage}{0.4\textwidth}
\centering
\vspace{0.2cm}
\includegraphics[width=0.9\textwidth]{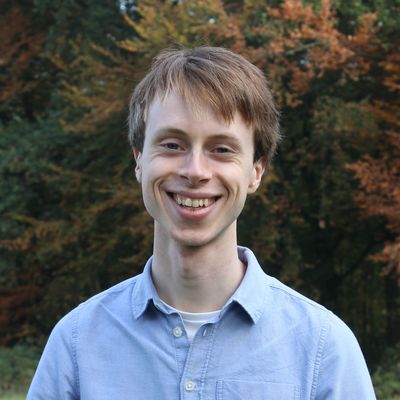}
\vspace{0.6cm}
\end{minipage}

%
%

\end{document}